\documentclass[conference]{IEEEtran}
\IEEEoverridecommandlockouts
\usepackage[utf8]{inputenc}

\usepackage{hyperref}
\usepackage{pbalance}
\usepackage{float}

\IEEEoverridecommandlockouts 
\usepackage[dvips]{graphicx}
\usepackage{algorithmic}
\usepackage{algorithm}
\usepackage[OT4,T1]{fontenc}
\usepackage[cmex10]{amsmath}
\usepackage{url}
\usepackage{multirow}

\title{Punctuation Prediction for Polish Texts using Transformers}
\author{
\IEEEauthorblockN{Jakub Pokrywka}
\IEEEauthorblockA{
Adam Mickiewicz University\\
Faculty of Mathematics and Computer Science,\\
Email: jakub.pokrywka@amu.edu.pl}
}

\begin{document}
\maketitle              

\begin{abstract}
Speech recognition systems typically output text lacking punctuation. However, punctuation is crucial for written text comprehension. To tackle this problem, Punctuation Prediction models are developed. This paper describes a solution for Poleval 2022 Task 1: Punctuation
Prediction for Polish Texts, which scores 71.44 Weighted F1. The method utilizes a single HerBERT model finetuned to the competition data and an external dataset.
\end{abstract}

\section{Introduction}

Automatic Speech Recognition (ASR) systems produce speech transcripts, which typically do not contain punctuation. This may negatively impact the overall clarity of the transcribed text.  For several reasons, punctuation is important:
\begin{itemize}
    \item Punctuation reduces ambiguity in communication. The Sentences "Let's eat, children" and "Let's eat children" have completely different meanings, but they only vary in a comma.
    \item Punctuation helps in clarifying the intended meaning of a text. It provides cues to understand the structure of the text. Punctuation marks like commas, periods, question marks, and exclamation marks indicate pauses, sentence endings, and changes in tone or intent.
    \item Punctuation conveys tone and emotion behind the text. E.g., an exclamation mark may indicate excitement and a question mark may denote uncertainty. 
    \item Punctuation enhances the readability of the written words. Breaking down complex sentences into smaller parts with the use of commas, colons, and semicolons creates pauses, which aids in understanding the text
\end{itemize}

Many post-processing steps may be taken to circumvent this problem and the lack of capitalization problem. Such tasks are:

\begin{itemize}
    \item Punctuation Restoration (PR)
    \item Punctuation Prediction (PP)
    \item Capitalization Restoration  (CR)
\end{itemize}

The task of Punctuation Restoration is defined as the act of reinstating the original punctuation found in read speech transcripts.

This work describes the solution to Poleval 2022 Task 1: Punctuation Prediction from conversational language. The solution is based on the HerBERT model \cite{herbert} fine-tuned to the competition data and an external dataset.

\section{Related Work}

In the previous PolEval edition, a task similar to Punctuation Prediction was assigned, precisely PolEval 2021 Task: Punctuation restoration from read text \cite{poleval21punct}. The challenge unveiled WikiPunct, a fresh collection of text and audio corpus comprising 39 hours of audio and approximately 38,000 text transcripts. Four submissions \cite{wrobel,ropiak,marcinczuk,zietkiewicz} applied transformer-based methods for token classification, from which two authors utilized ensembles. Additionally, one author explored the integration of a bi-LSTM layer at the top of the transformer, along with vectors acquired from a wave2vec model.

When it comes down to other languages, authors of \cite{attia2014gwu} developed a method on  Support Vector Machines with Conditional Random Field (CRF) classifiers, using part-of-speech (POS) and morphological data for Arabic texts. Authors of \cite{che2016punctuation} used Deep Neural Networks and Convolutional Neural Networks for English texts and authors of \cite{sunkara2020robust} used transformers for English medical texts.

Recently, The Sentence End and Punctuation Prediction for many languages shared task was launched \cite{nlg21}. All of the teams explored neural network models, particularly transformers. The winning team described their solution in \cite{fullstop}.

\section{Competition Description}

The three datasets are provided for in the competition: train, dev, and test. For each dataset, input audio WAV files with text transcribed by an ASR system are delivered. The input text is segmented, where a single space separates each word. Each word is prepended by a word start timestamp and word end timestamp in milliseconds.

The missing punctuation symbols are as in table \ref{tab:allpuncts}.
\begin{table}[h]
    \centering
    \caption{Punctuation symbols in the challenge.} 
\begin{tabular}{c|c}
    symbol description & symbol character \\
    Fullstop & . \\
    Comma & , \\
    Question Mark & ? \\
    Exclamation Mark & ! \\
    Hyphen & - \\
    Ellipsis & … \\
\end{tabular}
    \label{tab:allpuncts}
\end{table}

The competition dataset is based on three resources summarized in Table \ref{tab:datasetstats}.

\begin{table*}[h]
    \centering
    \caption{The full competition dataset (train, dev, test) statistics.} 
\begin{tabular}{c|c|c|c|c|c|c}
    Subset &  Corpus & Files & Words & Audio [s] & Speakers & License  \\
    \hline
    CBIZ \cite{cbiz}  &	DiaBiz & 	69 & 	36   250 & 	16   916 & 	14 & 	CC-BY-SA-NC-ND \\ 
    VC	& Video conversations &	8&	44 656&	17  123&	20&	CC-BY-NC \\
    Spokes \cite{spokes}	&Casual conversations&	13&	42 730&	20 583&	19&	CC-BY-NC \\
\end{tabular}
    \label{tab:datasetstats}
\end{table*}

The dataset is split into three subsets as described in Table \ref{tab:datasetsplit}.

\begin{table}[h]
    \centering
    \caption{Competition dataset statistics split into train, dev, test.} 
\begin{tabular}{c|c|c|c|c}
    Dataset & Files & Words & Audio [s]  & License  \\
    \hline
    Train	&69&	98 095&	44 030&	CC-BY-SA-NC-ND \\
    Dev	    &11&	12 563&	4 718&	CC-BY-NC \\
    Test	&10&	12 978&	5 874&	CC-BY-NC \\
\end{tabular}
    \label{tab:datasetsplit}
\end{table}

The annotation scheme is not publicly available during the competition and will described in \cite{evaluationprocedure}.

There is one sample data from the training dataset in the subsection below.
\subsection{Sample data}
\textbf{Input wav file} : audio/AU1\_P1\_w\_drodze\_do\_sklepu.wav

\textbf{Input text} : I:5880-5880 teraz:5940-6180 mamy:6330-6450 drugi:6480-6900 dzień:6960-7080 takiej:7170-7410 ładnej:7440-7650 pogody:7830-8400 Ała:8430-8430 Nie:8760-8820 bij:8850-8970 mnie:9120-9330 kijem:9450-9870 To:10020-10080 boli:10170-10260

\textbf{Golden truth} :  I teraz mamy drugi dzień takiej ładnej pogody... Ała! Nie bij mnie kijem! To boli!


\subsection{Utilized Data}
\label{data}
In our final solution, we did not use any audio data. Additionally, we decided not to include start and stop timestamps as we did not observe any significant improvement in their score after conducting multiple experiments. Throughout the training process, we experimented with four different sources.
\begin{itemize}
    \item Poleval 2022 Task 1: Punctuation Prediction from Conversational Language (this competition training dataset)
    \item Poleval 2021 Task 1: Punctuation Restoration from Read Text \cite{poleval21punct} (training dataset)
    \item Poleval 2021 Task 1: Punctuation Restoration from Read Text  (test dataset)
    \item europarl-v7.pl-en.pl \cite{europarl}
\end{itemize}

Regrettably, the europarl-v7.pl-en.pl dataset did not lead to a score improvement. Therefore, it was not utilized in our final solution.

We have carried out normalization procedures. Firstly, we transformed the text format from being split with timestamps to raw text format with timestamps included. Secondly, we replaced all three consecutive full stop characters "." (Unicode code: 81) with a single ellipsis character "…" (Unicode code: 8230). This modification was essential for utilizing the punctuation prediction library explained in Section \ref{method}. 

Table \ref{tab:trainingdatastsstats} presents the statistics for the training datasets used and competition final test data: test-B. Some punctuation marks are more popular than others, which is consequent in all the datasets. 
There are some differences between training and testing datasets, but they are insignificant. E.g., the Fullstop character is more common in the test-B dataset than in the train dataset (104.022 vs. 78.338). The same stays true for Comma (133.303 vs. 112.923). The PolEval 2022 dataset exhibits much more significant differences than the PolEval 2021 dataset. This is particularly evident in the Mean Words per Sample metric, as well as in most punctuation characters. While some characters like Fullstop, Comma, and Ellipsis are more prevalent in the PolEval 2022 dataset, Hyphen is less frequent, and the Exclamation mark remains relatively unchanged.

Below are samples of golden truths from each dataset, with the last two examples shortened.
\subsubsection{Sample Poleval 2022 Task 1 test-B sentence}
\emph{No dzień dobry pani. Tu mi się jakaś opłata za kartę pobrała.}
\subsubsection{Sample Poleval 2022 Task 1 train sentence}
\emph{I teraz mamy drugi dzień takiej ładnej pogody… Ała! Nie bij mnie kijem! To boli!}
\subsubsection{Sample Poleval 2021 Task1 train sentence}
\emph{
w wywiadzie dla " polski " jarosław kaczyński podkreślił, że informacje dotyczące radosława sikorskiego zagrażają interesowi państwa. " to naprawdę wszystko, co mogę na ten temat powiedzieć "- odpowiedział, gdy dziennikarz pytał o bardziej szczegółowe informacje. premier kaczyński sugeruje, że dobry kandydat po na szefa dyplomacji to np. jacek saryusz- wolski wymieniony polityk zyskał uznanie braci kaczyńskich za dotychczasową działalność w charakterze dyplomaty i dużą wiedzę. "}

\subsubsection{Sample Poleval 2021 Task1 test sentence}
\emph{
801 co znaczy, że beginki " padły ofiarą reformacji "? grzesie2k wpis na słabym poziomie bzdurna informacja o 50 spalonych waldensach; po co w bibliografii pseudonaukowa książka magdaleny ogórek? fragment recenzji z księgarni gandalf: "magdalena ogórek do inkwizycji oraz kościoła ma stosunek jednoznaczny, pisząc o inkwizycyjnej pożodze oraz występkach heretyków spreparowanych przez inkwizytorów, którzy siali spustoszenie oraz o tym jak to w połowie xiii w? duchowni skupiali się na obsadzaniu stanowisk kościelnych, budowaniu zamętu przez interdykty, schizmy i walki, lekceważyli obowiązki duszpasterskie. nie ukrywa też, że jej celem jest próba rehabilitacji heretyków. takie jednoznacznie ideologiczne ustawienie problematyki nie ma wiele wspólnego z prawdą o epoce, obiektywizmem historycznym.}

\begin{table*}[htp]

    \centering
    \caption{Datatasets Statistics. The number of punctuation symbols is normalized per 1000 words.} 
\begin{tabular}{c|r|r|r|r|r|r|r|r}
    Dataset & Samples & Mean Words per Sample &  Fullstop & Comma & Question Mark & Exclamation Mark & Hyphen & Ellipsis \\
    \hline
    Poleval 2022 Task1 test-B & 1642&7.90&104.022&133.303&18.493&0.848&0.154&33.981\\
    Poleval 2022 Task1 train &    10601&8.87&78.338&112.923&16.718&2.574&1.67&47.039 \\ 

    Poleval 2021 Task1 train  & 800&206.39&63.405&61.364&4.827&0.715&14.826&0.018 \\ 

    Poleval 2021 Task1 test   & 200&204.21&62.999&61.163&3.648&0.563&15.205&0.0 \\
    europarl-v7.pl-en.pl      &  632565&20.26&50.086&76.627&1.383&3.354&7.32&0.097 \\

\end{tabular}
    \label{tab:trainingdatastsstats}
    \bigskip
    
    \centering
    \caption{final testing dataset test-B scores.}  
\begin{tabular}{c|c|c|c|c|c|c|c}
    model  & Weighted-F1 & Fullstop-F1 & Comma-F1 & Question Mark-F1 & Exclamation Mark-F1 & Hyphen-F1&  Ellipsis-F1 \\
    \hline
    allegro-herbert-large-cased-pl & 71.44 & 78.67&  72.25 &  74.96 &  16.67 &100.00& 43.72 \\
    polish-roberta-pl & 66.23 & 74.56 & 68.31 & 72.77 & 28.57 &100.00& 29.86
\end{tabular}
    \label{tab:testbresults}

\bigskip
    \caption{preliminary testing dataset test-A scores.} 
    \begin{tabular}{c|c|c|c|c|c|c|c}
    model  & Weighted-F1 & Fullstop-F1 & Comma-F1 & Question Mark-F1 & Exclamation Mark-F1 & Hyphen-F1 & Ellipsis-F1 \\
    \hline
    allegro-herbert-large-cased-pl & 67.30 & 77.32 & 70.31 & 76.23 &6.2 &100.00&  38.20 \\
    polish-roberta-pl & 62.17 & 71.6 & 66.88 & 69.15 & 22.86 &100.00& 28.92 \\
\end{tabular}
    \label{tab:testaresults}
\end{table*}

\subsection{Metric}



The challenge metric is the Weighted F1 score. The evaluation script is implemented in the GEval evaluation tool \cite{geval}. The challenge was hosted on the gonito platform \cite{gonito}. The final evaluation is done on the test-B dataset on all the domains. The metric definition is meticulously described in Poleval 2021 Task1 summary paper \cite{poleval21punct}.

\section{Method}

Our method was based on FullStop: Multilingual Deep Models for Punctuation Prediction \cite{fullstop} library. We slightly modified the library to work on a different set of punctuation marks than it was intended to. The final solution model was based on a single HerBERT \cite{herbert}, a neural model of transformer architecture \cite{attentinisallyouneed} trained on a corpus of Polish texts. The model was finetuned to the data described in Section \ref{data} with the aforementioned text preprocessing steps. We used scripts available at \url{https://github.com/oliverguhr/fullstop-deep-punctuation-prediction/blob/main/other_languages/readme.md}. The Polish RoBERTa \cite{polishroberta} model was evaluated as well, but not used for the final solution due to worse results. Both evaluations are available in Tables \ref{tab:testbresults} and \ref{tab:testaresults}. We also conducted experiments with XLM-RoBERTa \cite{xlmroberta}, but unfortunately, we did not achieve better results again.

\label{method}

\section{Results}
The final model using achieved a third-place score of 71.44 in the competition's Weighted F1 category. While it falls behind the first-place score of 83.30 and the second-place score, it still surpasses the baseline score of 35.30. Frequent punctuation symbols like full stops and commas (occurring above ten times per 1000 words) consistently scored between 70 and 80 in F1. However, the F1 scores varied greatly for less frequent symbols, with scores of 16.67, 100.00, and 43.72.


The subsections below illustrate some correct and incorrect predictions from the test-B dataset.
\subsection{Correct predictions}

\textbf{Predicted}: Nie rozumiem powodu, dla którego komuś za ciężko jest rozbić jajko. \\  \

\textbf{Predicted}: A ty dasz radę zabrać to wszystko?

\subsection{Incorrect predictions}
\textbf{Expected}:Ona nie będzie już\textbf{,}

\textbf{Predicted}:Ona nie będzie już\textbf{...} \\ \ 

\textbf{Expected}:Stary d\textbf{-} delegacyjny sprzęt z czasów PRLu, ale może być przydatny.

\textbf{Predicted}:Stary d\textbf{,} delegacyjny sprzęt z czasów PRLu, ale może być przydatny. \\ \ 

\textbf{Expected}:Zamknęli nam łazienkę\textbf{...} dranie\textbf{...}

\textbf{Predicted}:Zamknęli nam łazienkę\textbf{,} dranie 

\section{Conclusions}
In this paper, we proposed our solution to Poleval 2022 Task 1: Punctuation Prediction for Polish Texts. The method uses a single HerBERT model fine-tuned to the competition training data and other external datasets. The achieved score is 71.44, which falls behind the two best solutions but is significantly better than a baseline.

\bibliography{bibliography}
\bibliographystyle{ieeetr}
\end{document}